\documentclass[conference]{IEEEtran}
\IEEEoverridecommandlockouts
\usepackage{cite}
\usepackage{amsmath,amssymb,amsfonts}
\usepackage{algorithmic}
\usepackage{graphicx}
\usepackage{hyperref}  
\usepackage{textcomp}
\usepackage{xcolor}
\usepackage{flushend} 
\def\BibTeX{{\rm B\kern-.05em{\sc i\kern-.025em b}\kern-.08em
    T\kern-.1667em\lower.7ex\hbox{E}\kern-.125emX}}
\newcommand\mdoubleplus{\mathbin{+\mkern-10mu+}}

\begin{document}

\title{Language-Agnostic Syllabification with Neural Sequence Labeling}

\author{\IEEEauthorblockN{Jacob Krantz}
\IEEEauthorblockA{\textit{Dept. of Computer Science} \\
\textit{Gonzaga University}\\
Spokane, WA, USA \\
jkrantz@zagmail.gonzaga.edu}
\and
\IEEEauthorblockN{Maxwell Dulin}
\IEEEauthorblockA{\textit{Dept. of Computer Science} \\
\textit{Gonzaga University}\\
Spokane, WA, USA \\
mdulin2@zagmail.gonzaga.edu}
\and
\IEEEauthorblockN{Paul De Palma}
\IEEEauthorblockA{\textit{Dept. of Computer Science} \\
\textit{Gonzaga University}\\
Spokane, WA, USA \\
depalma@gonzaga.edu}
}

\maketitle

\begin{abstract}
The identification of syllables within phonetic sequences is known as \textit{syllabification}. This task is thought to play an important role in natural language understanding, speech production, and the development of speech recognition systems. The concept of the syllable is cross-linguistic, though formal definitions are rarely agreed upon, even within a language. In response, data-driven syllabification methods have been developed to learn from syllabified examples. These methods often employ classical machine learning sequence labeling models. In recent years, recurrence-based neural networks have been shown to perform increasingly well for sequence labeling tasks such as named entity recognition (NER), part of speech (POS) tagging, and chunking. We present a novel approach to the syllabification problem which leverages modern neural network techniques. Our network is constructed with long short-term memory (LSTM) cells, a convolutional component, and a conditional random field (CRF) output layer. Existing syllabification approaches are rarely evaluated across multiple language families. To demonstrate cross-linguistic generalizability, we show that the network is competitive with state of the art systems in syllabifying English, Dutch, Italian, French, Manipuri, and Basque datasets.
\end{abstract}


\section{Introduction}
Words can be considered compositions of syllables, which in turn are compositions of phones. Phones are units of sound producible by the human vocal apparatus. Syllables play an important role in prosody and are influential components of natural language understanding, speech production, and speech recognition systems. Text-to-speech (TTS) systems can rely heavily on automatically syllabified phone sequences \cite{pradhan2013syllable}. One prominent example is Festival, an open source TTS system that relies on a syllabification algorithm to organize speech production \cite{taylor1998architecture}.

Linguists have recognized since the late 1940s that the syllable is a hierarchical structure, present in most, if not all, languages (though there is some disagreement on this score. See, for example, \cite{kohler1966syllable}). An optional consonant onset is followed by a rime, which may be further decomposed into a high sonority vowel nucleus followed by an optional consonant coda. All languages appear to have at least the single syllable vowel ($V$) and the two syllable vowel-consonant ($VC$) forms in their syllable inventories. For example, \textit{oh} and \textit{so} in English. Most languages supplement these with codas to form the $\{V, CV, VC, CVC\}$ syllable inventory. Sonority rises from the consonant onset to the vowel nucleus and falls toward the consonant coda, as in the English \textit{pig}.

The components of the syllable obey the phonotactic constraints of the language in which they occur, and therein lies the question that motivates this research. Phonologists agree that the human vocal apparatus produces speech sounds that form a sonority hierarchy, from highest to lowest: vowels, glides, liquids, nasals, and obstruents. Examples are, \textit{c\textbf{o}me}, \textit{t\textbf{w}ist}, \textit{\textbf{l}ack}, \textit{ri\textbf{ng}}, and \textit{\textbf{c}at}, respectively. English, and other languages with complex syllable inventories, supplement the basic forms in ways that are usually consistent with the sonority hierarchy, where \textit{usually} is the operative word. Thus, English permits double consonant onsets, as in \textit{twist} with a consonant lower in the hierarchy (\textit{t}, an obstruent) followed by a consonant one higher in the hierarchy (\textit{w}, a glide). So sonority rises to the vowel, \textit{i}, falls to the fricative, \textit{s}, an obstruent, and falls further to another obstruent, \textit{t}, still lower in the hierarchy. Yet \textit{p} and \textit{w} do not form a double consonant onset in English, probably because English avoids grouping sounds that use the same articulators, the lips, in this instance. Constructing an automatic syllabifier could be the process of encoding all rules such as these in the language under investigation. Another approach, one more congenial to the rising tide of so-called usage-based linguists (e.g, \cite{bybee2010language}), is to recognize that the regularities of language formulated as rules can be usefully expressed as probabilities \cite{de2010syllables,kenstowicz1994phonology, selkirk1982syllable}.

An automatic syllabifier is a computer program that, given a word as a sequence of phones, divides the word into its component syllables, where the syllables are legal in the language under investigation. Approaches take the form of dictionary-based look-up procedures, rule-based systems, data-driven systems, and hybrids thereof \cite{marchand2009automatic}. Dictionary look-ups are limited to phone sequences previously seen and thus cannot handle new vocabulary \cite{hlaing2014automatic}. Rule-based approaches can process previously unseen phone sequences by encoding linguistic knowledge. Formalized language-specific rules are developed by hand, necessarily accompanied by many exceptions, such as the one noted in the previous paragraph. An important example is the syllabification package \textit{tsylb}, developed at the National Institute of Standards and Technology (NIST), which is based on Daniel Kahn's 1979 MIT dissertation \cite{fisher1996tsylb, kahn2015syllable}. Language particularity is a stumbling block for rule-based and other formal approaches to language such as Optimality Theory (OT), however much they strive for universality. Thus, T.A. Hall argues that the OT approach to syllabification found in \cite{hall2006towards} is superior to previous OT research as well as to Kahn's rule-based work, because both postulate language-specific structures without cross-linguistic motivation. From Hall's perspective, previous systems do not capture important cross-linguistic features of the syllable. In a word, the earlier systems require kludges, an issue for both builders of automatic, language-agnostic syllabifiers and theoretical linguists like Hall.

Data-driven syllabification methods, like the one to be presented in this paper, have the potential to function across languages and to process new, out of dictionary words. For languages that have transcribed syllable data, data-driven approaches often outperform rule-based ones. \cite{bartlett2009syllabification} used a combined support vector machine (SVM) and hidden Markov model (HMM) to maximize the classification margin between a correct and incorrect syllable boundary. \cite{rogova2013automatic} used segmental conditional random fields (SCRF). The SCRF hybrid method statistically leveraged general principles of syllabification such as legality, sonority and maximal onset. Many other HMM-based labeling structures exist, such as evolved phonetic categorization and high order n-gram models with back-off \cite{krantz2018syllabification, demberg2006letter}. 

Data-driven models are evaluated by word accuracy against transcribed datasets. Commonly, only one language or languages of the same family are used. The CELEX lexical database from \cite{baayen1995celex} contains syllabifications of phone sequences for English, Dutch, and German. These three languages fall into the West Germanic language family, so the phonologies of each are closely related. Evaluating a model solely on these three languages, the approach taken in \cite{rogova2013automatic} and others, does not adequately test a model's generalized ability to learn diverse syllable structures.

In this paper, we present a neural network that can syllabify phone sequences without introducing any fixed principles or rules of syllabification. We show that this novel approach to syllabification is language-agnostic by evaluating it on datasets of six languages, five from two major language families, and one that appears to be unrelated to any existing language. 

\section{Method} \label{method}

\begin{figure}
  \centering
  \includegraphics[scale=0.65]{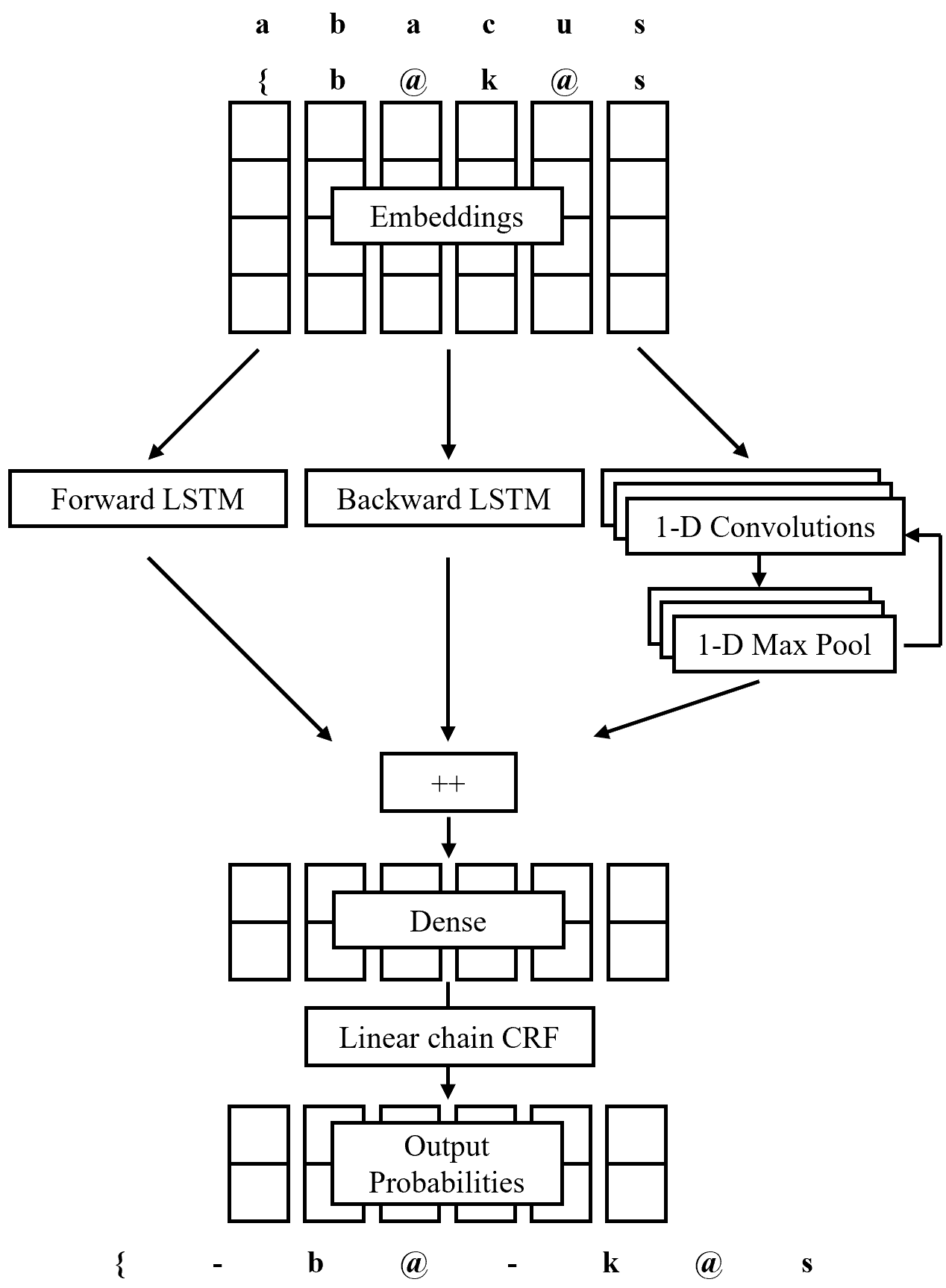}
  \caption{Network diagram detailing the concatenation of the forward and backward LSTMs with the convolutional component.}
  \label{fig:network}
\end{figure} 

Syllabification can be considered a sequence labeling task where each label delineates the existence or absence of a syllable boundary. As such, syllabification has much in common with well-researched topics such as part-of-speech tagging, named-entity recognition, and chunking \cite{wang2013effect}. Neural networks have recently outpaced more traditional methods in sequence labeling tasks. These neural-based approaches are taking the place of HMMs, maximum entropy Markov models (MEMM), and conditional random fields (CRF) \cite{young2018recent}.

In the following section and in Fig. \ref{fig:network}, we present a neural network architecture that leverages both recurrence and one-dimensional convolutions. Recurrence enables our model to read a sequence much like a human would; a sequence with elements $abcd$ would be read one element at a time, updating a latent understanding after reading each $a$, $b$, $c$, and finally $d$. One-dimensional convolutions extract a spatial relationship between sequential elements. The $abcd$ example sequence may then be read as $ab$, $bc$, $cd$. Explicitly recognizing this spatial relationship is beneficial in syllabification because a syllable is a local sub-sequence of phones within a word. The input to the model is a sequence of phones that together represent a word. We pad each phone sequence to a length of $n$ where $n$ is the length of the longest phone sequence. All inputs then take the form
\begin{equation}
    p=(p_0, p_1, ..., p_{n-2}, p_{n-1}).
\end{equation}
Each phone $p_i$ is mapped to a $d$-dimensional embedding vector $x_i$ resulting in
\begin{equation}
    x=(x_0, x_1, ..., x_{n-2}, x_{n-1})
\end{equation}
where $x$ has a dimension of $d\times n$. Taken together, the phone embeddings represent the relationships between phones in a real-valued vector space. The embedding dimension $d$ is optimized as a model hyperparameter and has a large impact on overall model performance \cite{reimers2017optimal}. As such, we carefully tune $d$ for the proposed \textit{Base} model and reduce it for our \textit{Small} model as described in Section \ref{results-section}.

The vector values of the phone embeddings are learned during each model training. Using learned embeddings enables the model to have a custom embedding space for each language that it is trained on. This is desirable because phonetic patterns differ from language to language. Also, learned embeddings allow the model to be trained using the input of any phonetic transcription. For example, one training of the model can use IPA and one can use SAMPA without needing to specify a mapping of one alphabet to another.

\subsection{Bidirectional LSTM}
\begin{figure}
  \centering
  \includegraphics[scale=0.26]{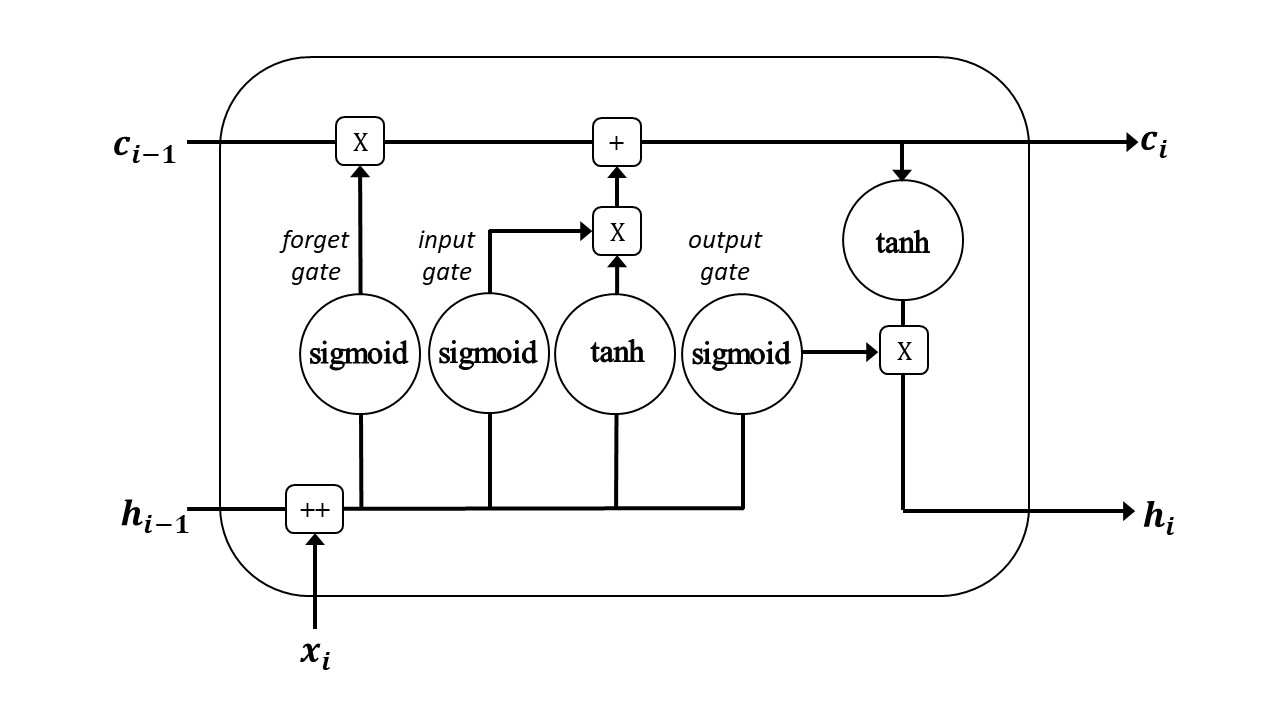}
  \caption{Diagram of the LSTM cell. $c_i$ and $h_i$ are the cell states and hidden states that propagate through time, respectively. $x_i$ is the input at time $i$ and is concatenated with the previous hidden state. $X$ represents element-wise multiplication and $+$ is element-wise addition. }
  \label{fig:lstm}
\end{figure}

Recurrent neural networks (RNNs) differ from standard feed-forward neural networks in their treatment of input order; each element is processed given the context of the input that came before. RNNs operate on sequential data and can take many forms. Our network leverages the long short-term memory (LSTM) cell which is a prominent RNN variant capable of capturing long-term sequential dependencies \cite{hochreiter1997long}. The gated memory cells of LSTM are an improvement over the standard RNN because the standard RNN is often biased toward short-term dependencies \cite{elman1990finding, bengio1994learning}. At each time step, the LSTM cell determines what information is important to introduce, to keep, and to output. This is done using an input gate, a forget gate, and an output gate shown in Fig. \ref{fig:lstm}. LSTM operates in a single direction through time. This can be a limitation when a time step has both past dependency and future dependency. For example, a consonant sound may be the coda of a syllable earlier in the sequence or the onset of a syllable later in the sequence. Thus, processing a phonetic sequence in both the forward and backwards directions provides an improved context for assigning syllable boundaries. A bidirectional LSTM (BiLSTM) is formed when an LSTM moving forward through time is concatenated with an LSTM moving backward through time \cite{graves2013speech}. 

We use the LSTM network as follows. The $x$ vector is fed through the LSTM network which outputs a vector $\overrightarrow{h_i}$ for each time step $i$ from $0$ to $n-1$. This is the forward LSTM. As we have access to the complete vector $x$, we can process a backward LSTM as well. This is done by computing a vector $\overleftarrow{h_i}$ for each time step $i$ from $n-1$ to $0$. Finally, we concatenate the backward LSTM with the forward LSTM: 
\begin{equation}
    h = [\overrightarrow{h}\mdoubleplus{}\overleftarrow{h}].
\end{equation}
Both $\overrightarrow{h_i}$ and $\overleftarrow{h_i}$ have a dimension of $l$, which is an optimized hyperparameter. The BiLSTM output $h$ thus has dimension $2l\times n$.

\subsection{CNN}
Convolutional neural networks (CNNs) are traditionally used in computer vision, but perform well in many text processing tasks that benefit from position-invariant abstractions \cite{lecun1998gradient, lopez2017deep}. These abstractions depend exclusively on local neighboring features rather than the position of features in a global structure. According to a comparative study by \cite{yin2017comparative},  BiLSTMs tend to outperform CNNs in sequential tasks such as POS tagging, but CNNs tend to outperform BiLSTMs in global relation detection tasks such as keyphrase matching for question answering. We use both the BiLSTM and the CNN in our network so that the strengths of each are incorporated. CNNs have been combined with BiLSTMs to perform state-of-the-art sequence tagging in both POS tagging and NER. \cite{ma2016end} used BiLSTMs to process the word sequence while each word's character sequence was processed with CNNs to provide a second representation. In textual syllabification, the only input is the phone sequence.

Both our BiLSTM and CNN components process the same input: the $x$ vector. We pad $x$ with $w-1$ $d$-dimensional zero vectors before $x_0$. A 1-dimensional convolutional filter of width $w$ processes a window $x_{i-w+1},...,x_i$ for all $i$ from $0$ to $n-1$. To determine the output vector $c$, the convolutional filter performs a nonlinear weight and bias computation. Due to the padding of $x$, the resulting dimension of $c$ is $f\times n$ where $f$ is the number of filters used. A 1-dimensional max pooling is performed over $c$ with a stride of 1 which keeps the dimensionality unaltered. The pool size is an optimized hyperparameter that determines how many adjacent elements are used in the $max$ operation. The convolutional and max pooling components can be repeated to compute higher-level abstractions. As the convolutional and max pooling output is conformant to the BiLSTM output, we can concatenate them to create a combined vector with dimension $(2l+f)\times n$:
\begin{equation}
    o = [h\mdoubleplus{}c].
\end{equation}

\subsection{Output: Conditional Random Field}
We introduce a time-distributed fully connected layer over vector $o$, taking $o$ from a dimension of $(2l+f)\times n$ down to a dimension of $2\times n$. We do this because there are two class labels: either a syllable boundary or no syllable boundary. The output of the model is a sequence
\begin{equation}
    y=(y_0,y_1,...,y_{n-2},y_{n-1}).
\end{equation}
When $y_i\equiv 0$, there is no syllable boundary predicted to follow the phone $p_i$. When $y_i\equiv 1$, there is a syllable boundary predicted to follow $p_i$. Intuitively, we seek an output sequence $y$ that gives the highest $p(y|o)$. One approach calculates the softmax for each $o_i$:
\begin{equation}
    s_i=\frac{e^{o_{i}}}{\sum_{k=0}^{1} e^{o_{i_k}}}.
\end{equation}
The softmax normalizes each $o_i$ to a probability distribution over the two discrete class labels. We can then model $p(y|o)$ by multiplying the maximum of each $s_i$ together:

\begin{equation}
    p(y|o)\approx\prod_{i=0}^{n-1} \max(s_i).
\end{equation}
When using the softmax, $p(y|o)$ is calculated under the limiting assumption that each $o_i$ is independent. To more accurately model $p(y|o)$, we replace the softmax classifier with a conditional random field (CRF) \cite{lafferty2001conditional}. Specifically, we use a linear-chain CRF which is a sequential model that leverages both past and future output tags to model the output probability. The linear-chain CRF can be considered a sequential generalization of logistic regression classifiers as well as a discriminative analogue of hidden Markov models because it models $p(y|o)$ directly instead of modeling $p(o|y)$ \cite{sutton2012introduction}. Using sequence-level tag information with a CRF has been shown to improve tag accuracy in the related tasks of POS tagging, chunking, and NER \cite{huang2015bidirectional,lample2016neural}. We use a linear-chain CRF to model the conditional distribution directly:
\begin{equation}
    p(y|o)\approx\frac{1}{Z(o)} \prod_{i=1}^{n-1} \exp \left \{ \sum_{k=1}^{K} \theta_{k}f_k(y_i,y_{i-1},o_i) \right\}
\end{equation}
where $Z(o)$ is the normalization function
\begin{equation}
    Z(o)= \sum_{y}^{} \prod_{i=1}^{n-1} \exp \left \{ \sum_{k=1}^{K} \theta_{k}f_k(y_i,y_{i-1},o_i) \right\}
\end{equation}
and $\theta$ is a learned parameter vector scaled by the set of transition feature functions $f$.

\subsection{Training}
Training of the network parameters is performed using backpropagation. Using Keras\footnote{\url{https://keras.io/}}, the backpropagation is automatically defined given the forward definition of the network. The defined loss function is sparse categorical cross entropy, in accordance with the real-valued probabilities given by the CRF output layer. Loss optimization is performed with the Adam optimizer \cite{kingma2014adam}. Adam was chosen because it adapts the learning rate on a parameter-to-parameter basis; strong convergence occurs at the end of optimization. Training is performed to a set number of epochs. Early stopping allows the network to conclude training if convergence is reached prior to reaching the epoch training limit \cite{caruana2001overfitting}.

\section{Materials}

\begin{table*}[t]
    \caption{Datasets and languages used for evaluation. Average phone and syllable counts are per word.}
    \centering
    \begin{tabular}{|c|c|c|r|c|r|r|}
    \hline
    \textbf{Language} & \textbf{Family Group}        & \textbf{Dataset}                         & \textbf{Word Count}   &\textbf{Encoding}                          & \textbf{Avg. Phone Count }    & \textbf{Avg. Syllable Count} \\ \hline\hline
    English           & Indo-European: West Germanic & CELEX \cite{baayen1995celex}             & 89,402                & DISC                    &  7.415                        & 1.788                           \\ \hline
    Dutch             & Indo-European: West Germanic & CELEX \cite{baayen1995celex}             & 327,548               & DISC                     &  8.469                        & 3.242                           \\ \hline
    Italian           & Indo-European: Romance       & Festival \cite{taylor1998architecture}   & 440,084               & SAMPA                       & 10.510                        & 3.513                           \\ \hline
    French            & Indo-European: Romance       & OpenLexique \cite{new2004lexique}        & 138,175               & Custom                     &  6.572                        & 1.870                           \\ \hline
    Manipuri          & Sino-Tibetan: Tibeto-Burman  & IIT-Guwahati \cite{singh2016automatic}   & 17,181                & Unknown                     &  7.258                        & 2.195                           \\ \hline
    Basque            & -                            & E-Hitz \cite{perea2006hitz}              & 100,079               & DISC                     &  8.979                        & 3.273                           \\ \hline
    \end{tabular}
    \label{tab:datasets}
\end{table*}

The materials for this research comprises the software described above and several syllabified datasets. 

\subsection{Software}
The implementation of our model was adapted from an open source code library\footnote{\url{https://github.com/UKPLab/emnlp2017-bilstm-cnn-crf}} designed for general-purpose sequence tagging and made available by \cite{reimers2017reporting}. The modifications to this code include adding data preparation scripts and changing the model architecture to reflect the network architecture described above. Our code is made publicly available for future research at \url{https://github.com/jacobkrantz/lstm-syllabify}.

\subsection{Datasets}

To produce a language-agnostic syllabifier, it is crucial to test syllabification accuracy across different language families and language groupings within families. We selected six evaluation languages: English, Dutch, Italian, French, Basque, and Manipuri. These represent two language families (Indo-European, Sino-Tibetan), a language isolate thought to be unrelated to any existing language (Basque), and two different subfamilies within the Indo-European family (West Germanic, Romance). The primary constraint was the availability of syllabified datasets for training and testing. Table \ref{tab:datasets} presents details of each dataset.

Among the six languages we evaluate with, both English and Dutch are notable for the availability of rich datasets of phonetic and syllabic transcriptions. These are found in the CELEX (Dutch Centre for Lexical Information) database \cite{baayen1995celex}. CELEX was built jointly by the University of Nijmegen, the Institute for Dutch Lexicology in Leiden, the Max Planck Institute for Psycholinguistics in Nijmegen, and the Institute for Perception Research in Eindhoven. CELEX is maintained by the  Max Planck Institute for Psycholinguistics. The CELEX database contains information on orthography, phonology, morphology, syntax and word frequency. It also contains syllabified words in Dutch and English transcribed using SAM-PA, CELEX, CPA, and DISC notations. The first three are variations of the International Phonetic Alphabet (IPA), in that each uses a standard ASCII character to represent each IPA character. DISC is different than the other three in that it maps a distinct ASCII character to each phone in the sound systems of Dutch, English, and German \cite{ldc2019guide}. Different phonetic transcriptions are used in different datasets. Part of the strength of our proposed syllabifier is that every transcription can be used as-is without any additional modification to the syllabifier or the input sequences. The other datasets were hand-syllabified by linguists with the exception of the IIT-Guwahat dataset and the Festival dataset. Both IIT-Guwahat and Festival were initially syllabified with a naive algorithm and then each entry was confirmed or corrected by hand.

For each dataset used to evaluate the proposed model, we compare our results with published accuracies of existing syllabification systems. Table \ref{tab:relatedwork} shows the performance of well known and state of the art syllabifiers for each dataset. Liang's hyphenation algorithm is commonly known for its usage in \TeX. The \textit{patgen} program was used to learn the rules of syllable boundaries \cite{adsett2008automatic}. What we call Entropy CRF is a method particular to Manipuri; a rule-based component estimates the entropy of phones and phone clusters while a data-driven CRF component treats syllabification as a sequence modeling task \cite{singh2016automatic}.

\begin{table}[t]
\caption{Reported accuracies of state of the art and selected high performing syllabifiers on each evaluation dataset.}
\centering
\begin{tabular}{|c|c|c|c|}
\hline
\textbf{Dataset} & \textbf{Syllabifier}                           & \textbf{Method} & \textbf{\%}   \\ \hline\hline
English CELEX    & \textit{tsylb} \cite{fisher1996tsylb}          & Rule-based      & 93.72         \\ \hline
English CELEX    & HMM-GA \cite{krantz2018syllabification}        & Data-driven     & 92.54         \\ \hline
English CELEX    & Learned EBG \cite{daelemans1992generalization} & Data-driven     & 97.78         \\ \hline
English CELEX    & SVM-HMM \cite{bartlett2009syllabification}     & Data-driven     & 98.86         \\ \hline
Dutch CELEX      & SVM-HMM \cite{bartlett2009syllabification}     & Data-driven     & 99.16         \\ \hline
Festival         & Liang hyphenation \cite{adsett2008automatic}   & Data-driven     & 99.73         \\ \hline
OpenLexique      & Liang hyphenation \cite{adsett2008automatic}   & Data-driven     & 99.21         \\ \hline
IIT-Guwahat      & Entropy CRF \cite{singh2016automatic}          & Hybrid          & 97.5          \\ \hline
E-Hitz           & Liang hyphenation \cite{adsett2008automatic}   & Data-driven     & 99.68         \\ \hline
\end{tabular}
\label{tab:relatedwork}
\end{table}

\section{Experiments}
Each dataset used to evaluate the model was split into three groups: training, development, and test. Each training epoch iterated over the training set to optimize the model parameters. The development set was used to tune the hyperparameters of the model, such as the batch size and the phone embedding dimension. The test set was exclusively used for reporting the model accuracy. The datasets were split randomly by percentages $80$ (training), $10$ (development), and $10$ (test). For the English CELEX dataset of $89,402$ words, this resulted in $71,522$ words for training and $8,940$ words for each development and training. 

For each experiment, models were initialized with a random set of parameter weights. \cite{reimers2017reporting} showed that differences in random number generation produce statistically significant variances in the accuracy of LSTM-based models. Due to the stochastic nature of neural network training, we performed each experiment $20$ times. We report model accuracy as a mean and standard deviation of these experiment repetitions.

\subsection{Data Cleaning}
Prior to splitting each dataset, a simple cleaning process had to be performed to remove unwanted entries. This cleaning involved removing all entries that had at least one other entry with the same word. It is important to note that two words being different does not necessitate a different pronunciation or syllabification. These entries with different words but same pronunciations were kept in the dataset. No other cleaning was needed for the datasets other than mapping the syllabified phone sequence to an input-target pair usable by our model for training and evaluation. This cleaning process contributes to the language-agnostic nature of this research. The simplicity of the cleaning process is enabled by the fact that the model is end to end; no external phonetic features are gathered, and any phonetic transcription can be accommodated in the training process.

\subsection{Hyperparameter Specification}
For all experiments, models were trained with a batch size of $64$. A limit of $120$ epochs was imposed with early stopping after $10$ unimproved epochs. Dropout was used for the input connection to the BiLSTM layer at $25\%$ \cite{srivastava2014dropout}. The learned embeddings layer had dimension $d=300$. The LSTM outputs, $\overrightarrow{h_i}$ and $\overleftarrow{h_i}$, both had dimension $l=300$. The convolutional to max pooling component was repeated twice before concatenation with the BiLSTM output. $200$ convolutional filters were used and each had a dimension of $3$. Finally, when using the Adam optimizer, we scaled the gradient norm when it exceeded $1.0$ using the Keras \textit{clipnorm} parameter. All training was performed on single GPU machines on Amazon Web Services (AWS) servers which provided more than enough compute power. The average training of a model on the English CELEX dataset took approximately $45$ minutes to reach convergence.

\subsection{Results} \label{results-section}

    \begin{table*}[t]
    \caption{The accuracy of our proposed model on each evaluation dataset. Model accuracy ($\% \pm\sigma$) is reported on a word level which means the entire word must be syllabified correctly.}
    \centering
    \begin{tabular}{|c|c|c|c|c|c|c|}
    \hline
    \textbf{Model}  & \textbf{English CELEX}    & \textbf{Dutch CELEX}      & \textbf{Festival}           & \textbf{OpenLexique}        & \textbf{IIT-Guwahat}      & \textbf{E-Hitz}           \\ \hline\hline
    \textbf{Base}   & \boldmath$98.5 \pm 0.1$   & \boldmath$99.47 \pm 0.04$ & \boldmath$99.990 \pm 0.005$ & $99.98 \pm 0.01$            & $94.9 \pm 0.3$            & \boldmath$99.83 \pm 0.07$ \\ \hline
    Small           & $98.2 \pm 0.2$            & $99.39 \pm 0.04$          & \boldmath$99.990 \pm 0.004$ & $99.987 \pm 0.007$          & \boldmath$95.4 \pm 0.3$   & $99.68 \pm 0.06$          \\ \hline
    Base-Softmax    & $97.7 \pm 0.2$            & $99.24 \pm 0.06$          & $99.984 \pm 0.003$          & \boldmath$100.00 \pm 0.01$  & $94.7 \pm 0.3$            & $99.71 \pm 0.04$          \\ \hline
    \end{tabular}
    \label{tab:ourresults}
    \end{table*}

We tested three model versions against all datasets. The model we call \textit{Base} is the BiLSTM-CNN-CRF model described in Section \ref{method} with the associated hyperparameters. Another model, \textit{Small}, uses the same architecture as \textit{Base} but reduces the number of convolutional layers to $1$, the convolutional filters to $40$, the LSTM dimension $l$ to $50$, and the phone embedding size $d$ to $100$. We also tested a \textit{Base-Softmax} model, which replaces the CRF output of the \textit{Base} model with a softmax. A comparison of the results of these three models can be seen in Table \ref{tab:ourresults}. This comparison empirically motivates the CRF output because \textit{Base} almost always outperforms \textit{Base-Softmax}. Of these three models, the \textit{Base} model performed the best with the exception of the French and Manipuri datasets. The differences in the French results can be considered negligible because the accuracies are all near $100\%$. The \textit{Small} model performed best on Manipuri, which may suggest that reducing the number of parameters of the \textit{Base} model leads to better accuracy on smaller datasets.

When comparing our model with previous syllabifiers, we consider the \textit{Base} model exclusively. In Table \ref{tab:english}, a side-by-side comparison of our \textit{Base} model to a selection of published syllabifiers shows that \textit{Base} is near state-of-the art performance on English CELEX. For the Dutch dataset, we report an accuracy of $99.47 \pm 0.04\%$, which improves on the previously best-known accuracy of $99.16\%$ from the HMM-SVM of \cite{bartlett2009syllabification}. Best-known results are also obtained on the Italian, French, and Basque datasets. Our reported accuracy of $94.9 \pm 0.3\%$ on the Manipuri dataset is furthest from state of the art. We suspect this to be due to having limited amounts of training data; the $97.5\%$ accurate system from \cite{singh2016automatic} supplemented their data-driven approach with rules of syllabification.

    \begin{table}[t]
    \caption{Comparison of reported accuracies against the English CELEX dataset. Note that HMM-SVM trained on 30K examples, Learned EBG trained on 60K, and HMM-GA trained on 54K.}
    \centering
    \begin{tabular}{|l|l|}
    \hline
    \textbf{Model}                                  & \textbf{Word Accuracy ($\% \pm\sigma$)}   \\ \hline\hline
    HMM-SVM \cite{bartlett2009syllabification}      & $98.86$                                   \\ \hline
    \textbf{BiLSTM-CNN-CRF (Base)(Ours)}            & $98.5 \pm 0.1$                            \\ \hline
    Learned EBG \cite{daelemans1992generalization}  & $97.78$                                   \\ \hline
    tsylb \cite{fisher1996tsylb}                    & $93.72$                                   \\ \hline
    HMM-GA \cite{krantz2018syllabification}         & $92.54$                                   \\ \hline
    \end{tabular}
    \label{tab:english}
    \end{table}

\section{Discussion}
    
    \begin{table*}[t] 
    \caption{Examples of generated syllabifications when the \textit{Base} BiLSTM-CNN-CRF model is trained on English CELEX. \textit{Target} is the syllabification given in English CELEX. Phones are represented in the DISC format and correct syllabifications are in bold.}
    \centering
    \begin{tabular}{|c|c|c|c|}
    \hline
    \textbf{Word}           & \textbf{Generated}             &   \textbf{Target}         \\ \hline\hline
    misinterpretation       &   \textbf{mIs-In-t3-prI-t1-SH} &   mIs-In-t3-prI-t1-SH     \\ \hline
    achieved                &   \textbf{@-Jivd}              &   @-Jivd                  \\ \hline
    worrisome               &  \textbf{wV-rI-sF}             &   wV-rI-sF                \\ \hline
    public-address systems  &   pV-blI-k-@-d-rEs-sI-st@mz    &   pV-blIk-@-drEs-sI-st@mz \\ \hline
    \end{tabular}
    \label{tab:examples}
    \end{table*}

Examples from the outputs of the \textit{Base} model can give us insight into what the model does well and what types of words it struggles with. The total number of sounds across languages is vast, but not infinite, as Ladefoged and Maddieson's \textit{The Sounds of the the World's Languages} demonstrates \cite{ladefoged1996sounds}. Different languages choose different inventories from the total producible by the human vocal apparatus. Within a language, sounds and patterns of sound vary widely in frequency, though with considerable regularity. This regularity has led a generation of linguists to attempt to uncover rules that describe not only syntax, but sound as well. Chomsky and Halle's \textit{The Sound Pattern of English} is the classic effort, first appearing in 1968 \cite{chomsky1968sound}. It is not surprising that the earliest attempts to produce automatic syllabifiers were based on just such rule collections. Nor is it surprising that the best-known rule-based syllabifier was inspired by a doctoral dissertation at MIT, Noam Chomsky's home institution for five decades. An alternative approach is to recognize that 1) rules can be reconceptualized as probabilities and 2) native speakers of a language have internalized those very probabilities. Nevertheless, where there is probability, there is ambiguity. With all of these caveats in mind, a few examples have been selected from our results to showcase the model as shown in Table \ref{tab:examples}.

The syllabification of \textit{misinterpretation} illustrates the model's ability to process longer words. Containing 14 phones and 5 syllables, this word demonstrates that the model's pattern finding technique works well regardless of the location of phonetic and syllabic patterns in the word. The model can accurately handle prefixes, correctly syllabifying \textit{mis-} as Table \ref{tab:examples} shows. Another word is \textit{achieved}. Inflected languages, such as English, use morphemes to distinguish mood, tense, case, and number, among others. Thus, the verb \textit{achieve} has several forms, or conjugates. The syllabifier correctly detected the stem and the past tense morpheme, \textit{ed}. An odd aspect of the English CELEX dataset is the occurrence of entries, $22,393$ of which, that either have hyphens or are multiple entirely separate words, such as \textit{public-address systems}. Because the phonetic representation does not denote hyphens or whitespace, the model has difficulties processing these words.

\section{Conclusion}
We proposed a sequential neural network model that is capable of syllabifying phonetic sequences. This model is independent of any hand-crafted linguistic knowledge. We showed that this model performs at or near state of the art levels on a variety of datasets sampled from two Indo-European, one Sino-Tibetan, and an apparently family-less language. Specifically, the proposed model achieved accuracies higher than any other we could find on datasets from Dutch, Italian, French, and Basque languages and close to the best-reported accuracy for English and Manipuri. Evaluating the performance of the syllabifier across diverse languages provides strong evidence that the proposed model is language-agnostic.

\subsection{Future Work}
With a language-agnostic syllabification system, any language can be syllabified given enough labeled training data. A problem is that many languages do not have large, labeled syllabification datasets. For example, we failed to find available and sufficient datasets in the Slavic languages of Russian and Serbian. This problem can be addressed either in a concentrated effort to create more labeled data or in the development of systems that require limited data.

\section*{Acknowledgment}
This research was supported in part by a Gonzaga University McDonald Work Award by Robert and Claire McDonald and an Amazon Web Services (AWS) grant through the Cloud Credits for Research program.

\bibliographystyle{./bibliography/IEEEtran}
\bibliography{./bibliography/IEEEabrv,./bibliography/ourbib}

\end{document}